\pdfoutput=1

\documentclass[11pt]{article}

\usepackage{acl}

\usepackage{times}
\usepackage{latexsym}

\usepackage[T1]{fontenc}

\usepackage[utf8]{inputenc}

\usepackage{microtype}

\usepackage{graphicx}
\usepackage{amsmath}
\usepackage{amssymb}
\usepackage{booktabs}
\usepackage{float}
\usepackage{array}
\usepackage{enumitem}
\usepackage{colortbl}
\usepackage{subcaption}
\usepackage{multirow}
\usepackage{xurl}

\newcommand{\loss}{\mathcal{L}}

\definecolor{Gray}{gray}{0.85}
\newcolumntype{a}{>{\columncolor{Gray}\centering}p{0.8cm}}

\title{CLIP4IDC: CLIP for Image Difference Captioning}

\author{Zixin Guo, Tzu-Jui Julius Wang, Jorma Laaksonen \\
  Department of Computer Science, Aalto University, Finland \\
  \texttt{\{zixin.guo, tzu-jui.wang, jorma.laaksonen\}@aalto.fi} \\}

\begin{document}
\maketitle
\begin{abstract}
Image Difference Captioning (IDC) aims at generating sentences to describe differences between two similar-looking images.
Conventional approaches learn an IDC model with a pre-trained and usually frozen visual feature extractor. 
Accordingly, two major issues may arise: (1) a large domain gap usually exists between the pre-training datasets used for training such a visual encoder and that of the downstream IDC task, and (2) the visual feature extractor, when separately encoding two images, often does not effectively encode the visual changes between two images.
Due to the excellent zero-shot performance of the recently proposed CLIP, we thus propose CLIP4IDC to transfer a CLIP model for the IDC task to address those issues. 
Different from directly fine-tuning CLIP to generate sentences, 
we introduce an adaptation training process to adapt CLIP's visual encoder to capture and align differences in image pairs based on the textual descriptions.
Experiments on three IDC benchmark datasets, CLEVR-Change, Spot-the-Diff, and Image-Editing-Request, demonstrate the effectiveness of CLIP4IDC.
\end{abstract}

\section{Introduction}

\label{sec:intro}
Tasks involving understanding and expressing visual contents are hard for machines because modelling relationships between the visual and textual domains requires sophisticated computational reasoning. 
As one of the tasks, image Captioning (IC)~\citep{vinyals2015show,xu2015show} aims at generating a coherent description given an image.
Extended from image captioning, Image Difference Captioning (IDC)~\citep{jhamtani2018learning,park2019robust} describes the subtle changes that appear in a pair of two similar images.
It is more challenging as a machine is required to recognize both visual objects and nuances in the pair.

\begin{figure}[t]
    \centering
    \begin{subfigure}[t]{0.88\linewidth}
        \centering
        \includegraphics[width=0.99\linewidth]{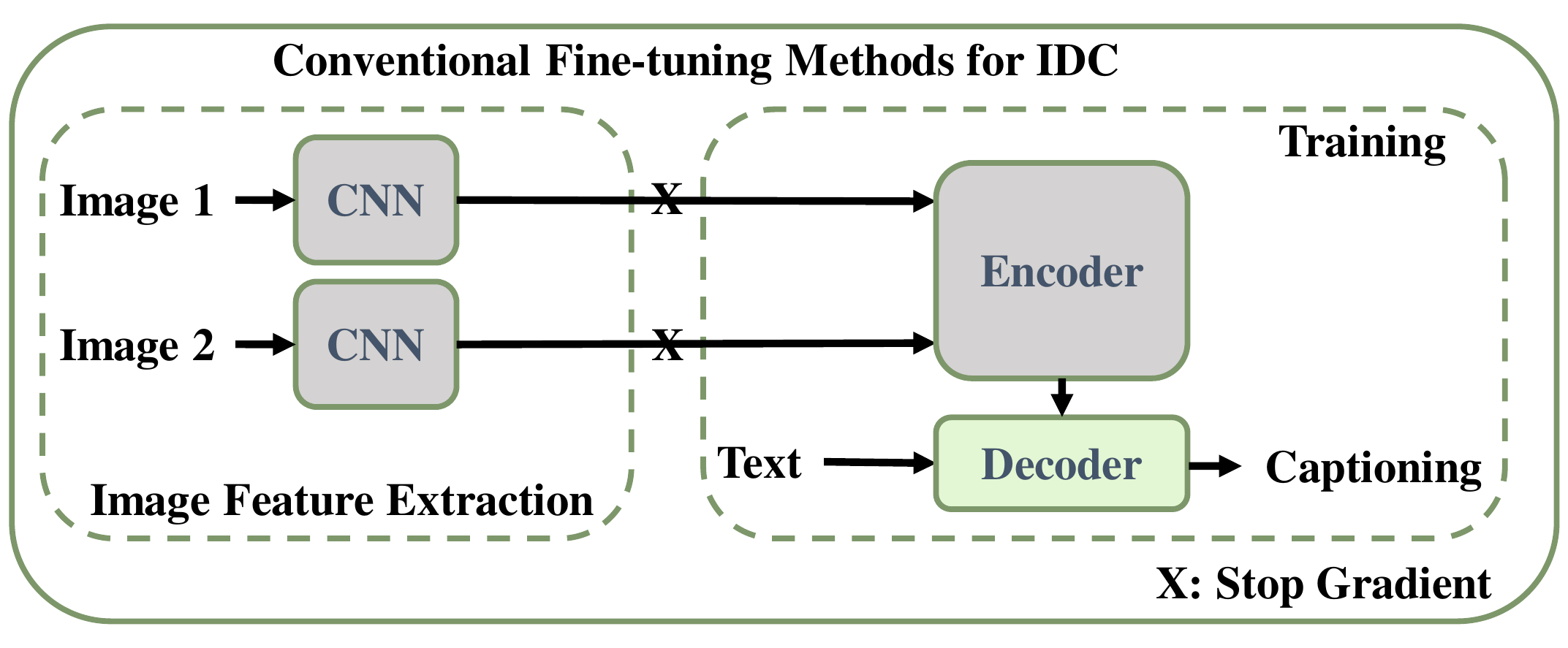}
        \caption{The fine-tuning strategy with a frozen (CNN) feature extractor. }
        \label{figure:frameworks}
    \end{subfigure}%
    
    \begin{subfigure}[t]{0.88\linewidth}
        \centering
        \includegraphics[width=\linewidth]{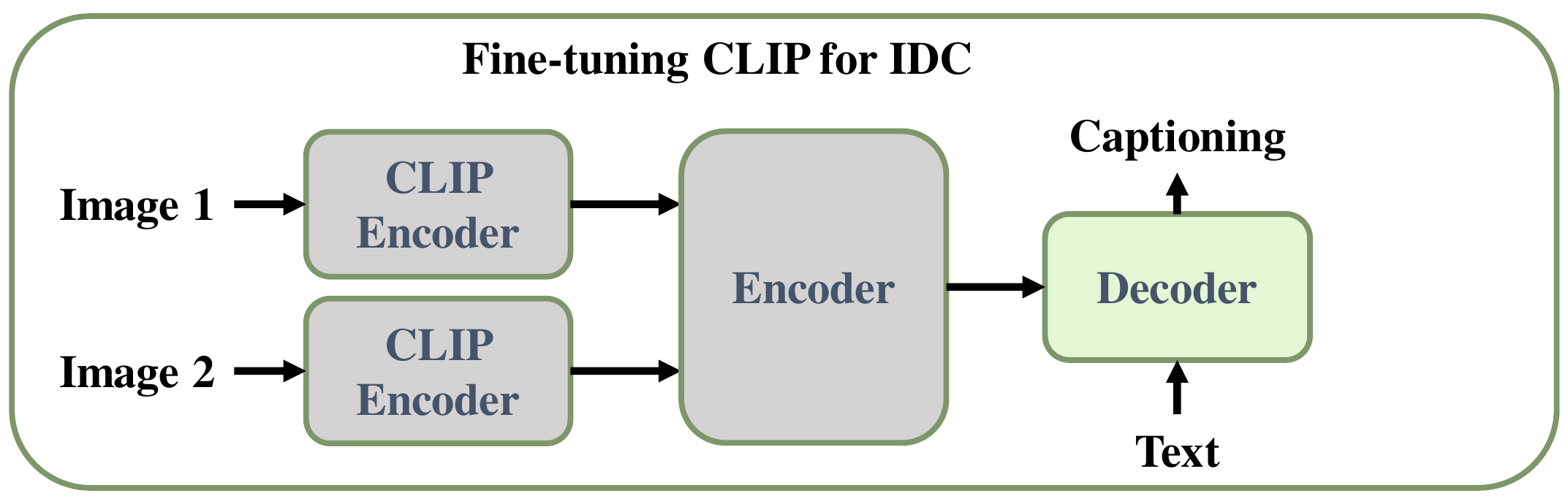}
        \caption{CLIP's fine-tuning strategy with an image encoder also fine-tuned.}
        \label{figure:finetune}
    \end{subfigure}
    \caption{Different conventional fine-tuning (FT) strategies may suffer from poor task accuracy due to: (1) not accounting for gaps introduced by either different objectives in pre-training (PT) and FT, and (2) domain shift in datasets used in PT and FT.}
    \vspace{-5mm}
\end{figure}

A conventional approach to IDC is shown in Figure~\ref{figure:frameworks}.
First, the visual features of an image pair are extracted offline with pre-trained models~\citep{he2016deep,ren2015faster}.
Then a captioning network generates sentence(s) to describe the changes in the pair. 
Even though such approaches have made great progress ~\citep{park2019robust,kim2021agnostic,huang2021image,hosseinzadeh2021image,sun2022bidirectional}, they suffer from the fact that the visual features do not account for the domain gap between the pre-training and IDC tasks.
\citet{lei2021less} demonstrated that the purpose of the feature extractor trained on the original task introduces a gap with that of the subsequent tasks. 
For example, the features extracted by models trained on image classification task focus on high-level context and lose fine-grained information required for IDC.
Moreover, the extracted visual representations of single modality are uncorrelated with the textual ones.

As an effective approach to deal with the drawbacks, fine-tuning models on the target dataset narrows the gap between the tasks.
\citet{yao2022image} showed that a Transformer~\cite{vaswani2017attention} model that was pre-trained and fine-tuned on the same offline-extracted features achieves state-of-the-art results in IDC. However, it does not yet fully exploit the knowledge from the large-scale dataset as in the recent advancements in vision-language (VL) pre-training~\cite{zhou2019unified,li2021align} (VLP).
In particular, CLIP~\cite{radford2021learning}, a contrastive VLP model has demonstrated its zero-shot superiority in numerous VL downstream tasks ~\cite{luo2021clip4clip,tang2021clip4caption}.

We set out experimenting with a typical CLIP fine-tuning strategy on the IDC task as shown in Figure~\ref{figure:finetune}, where CLIP's visual encoder is learned and fine-tuned on raw pixels. 
However, gaps still exist not only between the objectives of CLIP pre-training and IDC, but also between the collected image-text pairs for pre-training and the image difference pairs in IDC.
These gaps throttle the model in adapting for the IDC task.

\begin{figure*}[t]
\centering
\begin{minipage}{0.75\linewidth}
\centering
\includegraphics[width=\linewidth]{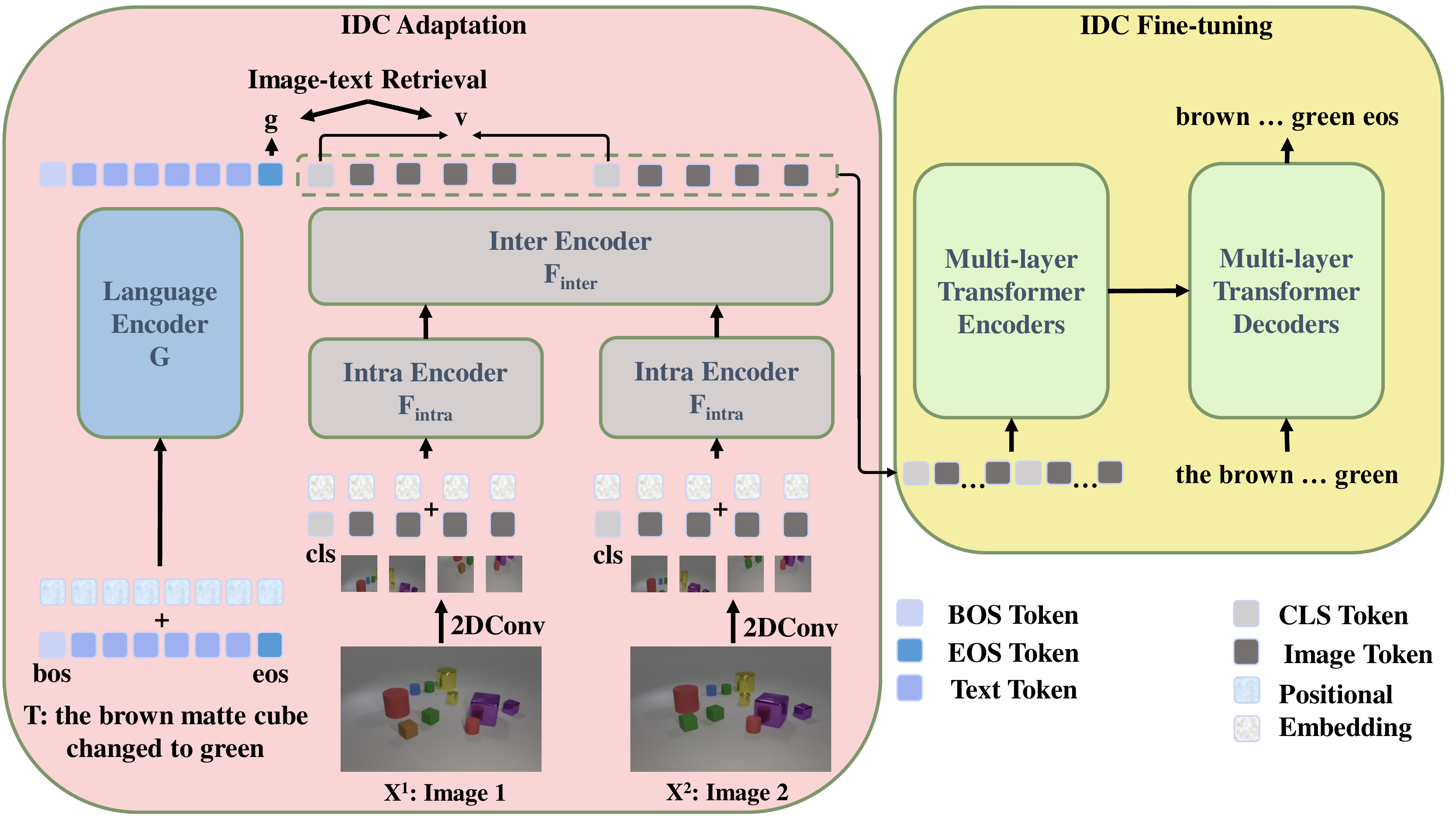}
\caption{The detailed architecture of CLIP4IDC.}
\label{figure:architecture}

\end{minipage}
\vspace{-5mm}

\end{figure*}

To tackle these problems, we study how to efficiently transfer a pre-trained CLIP for IDC. 
The overview of the proposed CLIP4IDC model is shown in Figure~\ref{figure:architecture}.
Compared to directly fine-tuning CLIP for the IDC task, CLIP4IDC employs "\textit{adapt}-and-\textit{fine-tune}" strategy. 
To \textit{adapt}, the CLIP encoder learns to capture the fine-grained differences in the image pair rather than to produce only high-level semantic information separately for these two images.
The visual and textual representations for the image pairs and the sentences are learned to be aligned with a retrieval loss in this stage. 
To \textit{fine-tune}, the learned vision encoder is followed by a captioning Transformer trained from scratch.

Extensive experiments are conducted on synthetic and real benchmark datasets CLEVR-Change~\cite{park2019robust} and Spot-the-Diff~\cite{jhamtani2018learning}, respectively. 
In addition, results on Image-Editing-Request~\cite{tan2019expressing}, a mixed real--synthetic dataset, are also reported.
CLIP4IDC outperforms the strong baselines on all the metrics on these three datasets. 
The main contributions of this work are:

 1) Compared with the conventional approaches that are trained on pre-extracted features, we fine-tune CLIP for IDC on raw pixels. This retains the expressiveness of the pre-trained features as well as adapting them to the new task domain.

 2) We propose CLIP4IDC, which consists of adaptation and fine-tuning stages, to narrow the gap between the objectives and data domains during pre-training CLIP and fine-tuning it for IDC. 
 The adaptation is learned by mutually retrieving the visual differences and the descriptions.

 3) Extensive experiments show that CLIP4IDC outperforms multiple strong baselines in the IDC task on three datasets on all the metrics.\footnote{\url{https://github.com/sushizixin/CLIP4IDC} }

\section{CLIP4IDC}
As shown in Figure~\ref{figure:frameworks}, 
the canonical IDC approach generates sentences on pre-extracted features.
The bottleneck lies in three aspects: 1) the stopped gradient flow in the feature extraction, 2) the mismatched objectives and data domains between the pre-training and IDC fine-tuning,
and 3) 
the visual features being "purely visual", i.e. they reside in the visual domain, far apart from the textual domain.
In the following sections, we introduce CLIP4IDC, a CLIP-based approach to address these bottlenecks. 

\subsection{CLIP Fine-tuning Approach}

An end-to-end approach of fine-tuning CLIP for IDC is shown in Figure~\ref{figure:finetune}. 
Specifically, the image representations are generated by the vision encoder initialized with CLIP~\cite{dosovitskiy2020image} and are fed into a Transformer encoder to focus on accounting for the differences in the image pair.
A Transformer decoder is applied to describe the changes given the visual context.

\subsection{Model Architecture}
Figure~\ref{figure:architecture} sketches the CLIP4IDC model, containing the vision and language encoders.

\noindent\textbf{Language Encoder.}
Given a textual caption $T$, the language encoder $G$ consisting of $N_G$ Transformer layers is used, denoted as:
\setlength{\abovedisplayskip}{2pt}
\setlength{\belowdisplayskip}{\abovedisplayskip}
\begin{equation}
G(T)=G(\{E_{\text{bos}},E_{t_1},...,E_{t_m},E_{\text{eos}}\}+p_{\text{T}}), 
\end{equation}
where $E_{*}\in \mathbb{R}^{d_T}$ is a linear projection of each token and $p_T\in\mathbb{R}^{(m+2)\times d_T}$
is a learned positional embedding to retain the positional information.
$E_{bos}$ and $E_{eos}$ are token embeddings to represent the start and end of the text, respectively.
The language encoder's output $g\in\mathbb{R}^{d_T}$ is generated by collecting the output of the token embedding $E_{eos}$.

\noindent\textbf{Vision Encoder.} 
Each image in the image pair $(X^{1}, X^{2})$ is patchified with the CLIP's initial convolutional layer into $n$ image patches with dimensionality $d_I$ as:
\setlength{\abovedisplayskip}{2pt}
\setlength{\belowdisplayskip}{\abovedisplayskip}
\begin{flalign}
X^{1}&=\{x_{\text{cls}},x_1^1,...,x_{n}^1\}+p_{\text{I}}, \\
X^{2}&=\{x_{\text{cls}},x_1^2,...,x_{n}^2\}+p_{\text{I}},
\end{flalign}
where $x_{cls}$ is a learned class embedding to represent the global context of the images and the positional embedding $p_{\text{I}} \in \mathbb{R}^{(n+1)\times d_I}$.
$\{\cdots\}$ is the sequence of the embeddings.
The vision encoder $F$ is constructed to capture the subtle changes in the image pair. $F$ is initialized by CLIP's weights and composed of a \emph{intra} and \emph{inter} Transformer modules.
Specifically, the \emph{intra} module $F_{intra}$ containing $N_{intra}$ Transformer layers learns the uni-modal context from the image pairs.
The \emph{inter} module $F_{inter}$ with $N_{inter}$ layers is constructed to focus on the subtle difference between the contexts in each pair.
These procedures are formulated as:
\begin{flalign}
F(X^1,X^2)=F_{inter}(\{&F_{intra}(X^1)+e_1,  \\ &F_{intra}(X^2)+e_2\}+p), \nonumber
\end{flalign}
where $p\in\mathbb{R}^{2(n+1)\times d_I}$. 
$e_1$ and $e_2\in\mathbb{R}^{d_I}$ are special token embeddings to represent the first and second images.
Afterwards, a learnable linear projection $W\in\mathbb{R}^{d_I\times d_T}$ is applied to the visual representation $F(X^1,X^2)$, on which the final visual representation $F'(X^1,X^2)$ is generated.

\subsection{IDC-specific Adaptation}
Next, we propose two novel IDC-specific pretext tasks, which are image-pair-to-text (IP-T) and text-to-image-pair (T-IP) retrieval, for better adapting the visual representations for captioning.

Prior to fine-tuning CLIP for the actual IDC task, we adapt the visual features to the domain of the IDC task via IP-T and T-IP retrieval. Our adaptation methodology follows the contrastive approach, where the encoded image pairs are drawn closer to the encoded difference captions. Although other kinds of adaptation strategies exist, such as the one focusing more on matching the domain distributions \cite{tzeng2014deep}, we only focus on testifying if adding such an adaptation step is useful.
We aggregate a combined visual representation $v\in\mathbb{R}^{d_T}$ of the image pair from their $x_{cls}$ embeddings, denoted as:
\begin{flalign}
v=f(\{F'(X^1,X^2)_1, F'(X^1,X^2)_{n+2}\}),
\end{flalign}
where $f$ is the mean-pooling operation. The subscript is the position (1-indexed) of the embeddings in the representation. 
\newcommand{\simx}{\text{s}}%
Given $B$ image pairs and difference captions in a batch, the target is to match $B\times B$ similarities between the difference representations of the image pairs and the descriptions to the differences.
The loss function is defined as:
\begin{flalign}
\loss_{i2t}&=\frac{-1}{B}\sum\limits_{i}^B \log\frac{\exp(\simx(v_i,g_i)/\tau)}{\sum_{j=1}^B \exp(\simx(v_i,g_j)/\tau)},\\
\loss_{t2i}&=\frac{-1}{B}\sum\limits_{i}^B \log\frac{\exp( \simx(v_i,g_i)/\tau))}{\sum_{j=1}^B \exp(\simx(v_j,g_i)/\tau)},\\
\loss&=\loss_{i2t}+\loss_{t2i}, \label{eq:retrieval_loss}
\end{flalign}
where $\loss_{i2t}$ and $\loss_{t2i}$ are the loss functions of IP-T and T-IP retrieval, respectively.
$\simx(\cdot,\cdot)$ denotes the cosine similarity function and $\tau$ is a learnable temperature parameter to smooth the gradients.

\vspace{-4pt}
\subsection{Captioning}
In the actual captioning stage, the vision encoder is initialized with the weights obtained from the previous adaptation stage and the output $F'(X^1,X^2)$ of the vision encoder is fed into the captioning model.
As shown in Figure~\ref{figure:architecture}, the captioning model contains multi-layer Transformer encoders and decoders for the visual and textual representations, respectively.
The decoder is trained to predict the next token given the previous ground truth words and the visual differences.
A word-level cross entropy (XE) loss as in~\citet{park2019robust} is utilized.

\section{Experiments}
\subsection{Benchmark Datasets and Metrics}
\noindent We conduct experiments on CLEVR-Change~\cite{park2019robust}, Spot-the-Diff~\cite{jhamtani2018learning} and Image-Editing-Request~\cite{tan2019expressing} datasets.
Following previous works, e.g.~\cite{huang2021image,hosseinzadeh2021image}, captioning models are evaluated on BLEU (B)~\cite{papineni2002bleu}, METEOR (M)~\cite{banerjee2005meteor}, CIDEr-D (C)~\cite{vedantam2015cider} and ROUGE-L (R)~\cite{lin-2004-rouge} on the $\textit{test}$ split.
IDC adaptation is done via image-pair-to-text (IP-T) and text-to-image-pair (T-IP) retrieval tasks.
The standard retrieval metrics are reported: recall at rank K (R@K), median rank (MdR) and mean rank (MnR).

\begin{table}[t]
\begin{minipage}{\linewidth}

\begin{center}

\scalebox{0.6}{\setlength\tabcolsep{6pt}
\begin{tabular}{p{4.1cm}
p{2.0cm}<{\centering} p{0.5cm}<{\centering}
p{0.6cm}<{\centering} p{0.6cm}<{\centering} 
a p{0.6cm}<{\centering}}
\toprule
Model & Input & PT &  B  &  M  & C &   R   \\
\midrule
Capt-Dual-Att~\citeyearpar{park2019robust} &  ResNet & --  & 43.5  &  32.7 & 108.5  &  -- \\
DUDA~\citeyearpar{park2019robust} &  ResNet &  --  & 47.3  &  33.9 & 112.0 &  -- \\
VAM~\citeyearpar{shi2020finding} &  ResNet  &   --  & 50.3  &  37.0 &  114.9 & 69.7 \\
VAM+~\citeyearpar{shi2020finding}  & ResNet  &   --  & 51.3  &  37.8 & 115.8  & 70.4 \\
IFDC~\citeyearpar{huang2021image}  & F-RCNN  &   --  & 49.2  &  32.5 & 118.7  &  69.1 \\
DUDA+Aux~\citeyearpar{hosseinzadeh2021image}  & ResNet  &  --  & 51.2  &  37.7 & 115.4  & 70.5  \\
VACC~\citeyearpar{kim2021agnostic} &  ResNet   &  -- & 52.4  & 37.5  &   114.2  & --  \\
BiDiff~\citeyearpar{sun2022bidirectional} &  ResNet   &   --  & \underline{54.2}  & \underline{38.3}  & 118.1 &  -- \\
IDC-PCL~\citeyearpar{yao2022image} &   ResNet   &   \checkmark  &  51.2 & 36.2  & \underline{128.9} & \underline{71.7} \\
CLIP4IDC   &   Raw   &  \checkmark  & \textbf{56.9}  & \textbf{38.4}  & \textbf{150.7} &  \textbf{76.4}   \\
\midrule
CC-Full~\citeyearpar{ak2022learning}  &  Raw,ResNet  &  \checkmark  &  64.3 & 36.4  & 151.4 &  77.1 \\
\bottomrule
\end{tabular}}
\end{center}
\vspace{-4mm}
\caption{Results of IDC 
on CLEVR-Change test split. The main metric CIDer is highlighted. CC-Full is in a separate group as it adopts the policy gradient method directly optimized for the target metrics.}
\label{table:clevr_results}

\end{minipage}
\vspace{-4mm}

\end{table}

\subsection{Captioning Results}
We compare CLIP4IDC against the direct CLIP fine-tuning method and the state of the arts which employ the pre-extracted features in Tables~\ref{table:clevr_results}--~\ref{table:edit_results}. 

\noindent\textbf{Results on CLEVR-Change.}
Table~\ref{table:clevr_results} shows that CLIP4IDC outperforms all the baselines except CC-Full~\cite{ak2022learning} on CIDEr.
Note that CC-Full employs the policy gradient method and is directly optimized for generating the target captions, while our proposed CLIP4IDC only relies on standard XE captioning loss. As such, we do not think their results are comparable, however, our results are still rather competitive. As we will see in a later section, CLIP4IDC significantly outperforms CC-Full on a real-world dataset.


\begin{table}[t]
\begin{minipage}{\linewidth}

\begin{center}

\scalebox{0.7}{\setlength\tabcolsep{7pt}
\begin{tabular}{p{3cm}
p{0.6cm}<{\centering}
p{0.8cm}<{\centering} p{0.6cm}<{\centering} p{0.6cm}<{\centering} p{0.6cm}<{\centering}
p{0.6cm}<{\centering}}
\toprule
Model &  C &  T  &  M  & A &   D & DI   \\
\midrule
DUDA~\citeyearpar{park2019robust} & 120.4  & 86.7  &  56.4 & 108.2 & 103.4 & 110.8  \\
VAM+~\citeyearpar{shi2020finding}  &  122.1  &  98.7 & 82.0  & 126.3 & 115.8  & 122.6  \\
IFDC~\citeyearpar{huang2021image}  &  \underline{133.2} & 99.1  &  \underline{82.1} & 128.2 & 118.5 & 114.2  \\
DUDA+Aux~\citeyearpar{hosseinzadeh2021image}  & 120.8   & 89.9  & 62.1  & 119.8 & 123.4 &  116.3  \\
BiDiff~\citeyearpar{sun2022bidirectional} &  115.9  & \underline{106.8}  &  71.8 & 121.3  & \underline{124.9} & 116.1   \\
IDC-PCL~\citeyearpar{yao2022image} & 131.2  & 101.1  &  81.7 & \textbf{133.3} & 116.5  & \textbf{145.0}   \\
CLIP4IDC  &  \textbf{149.1} &  \textbf{135.3} & \textbf{91.0} & \underline{132.4}  &  \textbf{135.5}  &  \underline{133.4} \\
\bottomrule
\end{tabular}}
\end{center}
\vspace{-4mm}

\caption{The breakdown of CIDEr score on different types of changes on CLEVR-Change test split. 
The columns C, T, M, A, D, DI stand for change types of Color, Texture, Move, Add, Drop and Distractor, i.e. no changes in the image pairs.
}
\label{table:type_results}

\end{minipage}

\end{table}

\begin{table}[t]
\begin{minipage}{\linewidth}

\begin{center}
\scalebox{0.6}{\setlength\tabcolsep{7pt}
\begin{tabular}{p{4.1cm}
p{2.0cm}<{\centering}
p{0.5cm}<{\centering}
p{0.6cm}<{\raggedleft} p{0.6cm}<{\centering} 
a p{0.6cm}<{\centering}}
\toprule
Model & Input & PT &  B  &  M  & C &   R   \\
\midrule
DDLA~\citeyearpar{jhamtani2018learning} & ResNet & --  & 8.5  &  12.0  & 32.8  &  28.6 \\
DUDA~\citeyearpar{park2019robust} &  ResNet  & --  & 8.1  &  11.5 & 34.0  &  28.3 \\
VAM~\citeyearpar{shi2020finding} & ResNet  & --  & \underline{10.1}  & 12.4  & 38.1  & 31.3  \\
IFDC~\citeyearpar{huang2021image}  &  F-RCNN & --  & 8.7  &  11.7 & 37.0  &  30.2 \\
DUDA+Aux~\citeyearpar{hosseinzadeh2021image} & ResNet   & -- & 8.1  & 12.5  & 34.5  &  29.9  \\
VACC~\citeyearpar{kim2021agnostic} &  ResNet  & --  &  9.7 & \underline{12.6}  &  \underline{41.5} &  \underline{32.1}  \\
CLIP4IDC &   Raw  & \checkmark & \textbf{11.6}  & \textbf{14.2}  &  \textbf{47.4} &   \textbf{35.0} \\
\midrule
CC-Full~\citeyearpar{ak2022learning} &  Raw,ResNet &   \checkmark  & 8.3  & 13.0 &  33.0 &  30.0 \\
\bottomrule
\end{tabular}}
\end{center}
\vspace{-4mm}
\caption{Results of IDC on Spot-the-Diff test split. }

\label{table:spot_results}

\end{minipage}

\end{table}

\begin{table}[h!]
\begin{minipage}{\linewidth}

\begin{center}
\scalebox{0.6}{\setlength\tabcolsep{7pt}
\begin{tabular}{p{4.1cm}
p{2.0cm}<{\centering} p{0.5cm}<{\centering}
p{0.6cm}<{\centering} p{0.6cm}<{\centering} 
a p{0.6cm}<{\centering}}
\toprule
Model & Input & PT & B  &  M  & C &   R   \\
\midrule
Rel-Att~\citeyearpar{tan2019expressing}  & ResNet  &  --   & 6.7  & 12.8  & 26.4  &  37.4 \\
DUDA~\citeyearpar{park2019robust} & ResNet  &  --  & 6.5  & 12.4  &  22.8 & 37.3  \\
BiDiff~\citeyearpar{sun2022bidirectional} & ResNet  &   --   & \underline{6.9}  & \textbf{14.6}  &  \underline{27.7} &  \underline{38.5} \\
CLIP4IDC  & Raw &  \checkmark  & \textbf{8.2}  & \textbf{14.6}  & \textbf{32.2}  &  \textbf{40.4}  \\
\bottomrule
\end{tabular}}
\end{center}
\vspace{-4mm}
\caption{Results on Image-Editing-Request test split. }
\label{table:edit_results}

\end{minipage}

\end{table}

\begin{table}[h!]
\begin{minipage}{\linewidth}

\begin{center}
\scalebox{0.6}{\setlength\tabcolsep{5.5pt}
\begin{tabular}{p{1.5cm}
p{0.2cm}<{\centering}
p{1.3cm}<{\centering}
p{0.5cm}<{\centering} p{0.5cm}<{\centering} a p{0.5cm}<{\centering}
p{0.5cm}<{\centering} p{0.5cm}<{\centering} a p{0.5cm}<{\centering}}
\toprule
  &   &  &  \multicolumn{4}{c}{CLEVR-Change}  & \multicolumn{4}{c}{Spot-the-Diff}  \\
Model & $\loss$  & Params  &  B  &  M  & C &  R &  B  &  M  & C &  R  \\
\midrule
CLIP-FT  &  --  &  135.57M & 49.9   & 34.8  & 133.9  & 70.8 &  11.0 & 12.8 & \underline{43.3} & \underline{33.5}  \\
CLIP4IDC  &  -- &   135.65M  &  \underline{54.2}  &  \underline{37.9} & \underline{147.5}  & \underline{75.4} & \underline{11.0}  & \underline{12.9} & 43.0 & 33.4   \\
CLIP4IDC  &  \checkmark  &  135.65M & \textbf{56.9} & \textbf{38.4}  & \textbf{150.7}  & \textbf{76.4}  & \textbf{11.6}  & \textbf{14.2}  & \textbf{47.4}  & \textbf{35.0} \\
\bottomrule
\end{tabular}}
\end{center}
\vspace{-4mm}
\caption{Ablation results of IDC on the two datasets. }
\label{table:ablation_results}

\end{minipage}
\vspace{-4mm}
\end{table}

We also assess the models by different types of changes on CLEVR-Change, as seen in Table~
\ref{table:type_results}. CLIP4IDC outperforms IDC-PCL on Color, Texture, Move and Drop types.

\noindent\textbf{Results on Spot-the-Diff and Image-Editing-Request.}
Tables~\ref{table:spot_results} and~\ref{table:edit_results} show that CLIP4IDC achieves higher accuracy than the baselines on all the metrics on the two real datasets.

\noindent\textbf{Ablations.}
We conduct ablation studies on different CLIP architectures and adaptation strategies.
Table~\ref{table:ablation_results} shows that CLIP4IDC without the adaptation stage (without $\mathcal{L}$ in Eq. \ref{eq:retrieval_loss}) outperforms the direct CLIP finetuning ("CLIP-FT") on CLEVR-Change. 
On the more challenging real-world dataset, Spot-the-Diff, we observe the same trend. 
Having the adaptation stage with $\mathcal{L}$ thus further enhances the performances. This confirms that learning to capture more fine-grained visual differences in the adaptation stage is beneficial.

\begin{table*}[htbp]
\begin{minipage}{\linewidth}

\begin{center}

\scalebox{0.7}{\setlength\tabcolsep{6pt}
\begin{tabular}{p{1.5cm}
p{0.6cm}<{\centering}
p{0.6cm}<{\centering} p{0.6cm}<{\centering}
p{0.6cm}<{\centering}
p{0.6cm}<{\centering} p{0.6cm}<{\centering}
p{0.6cm}<{\centering} p{0.6cm}<{\centering}
p{0.6cm}<{\centering}
p{0.6cm}<{\centering} p{0.6cm}<{\centering}
p{0.6cm}<{\centering}
p{0.6cm}<{\centering}
p{0.6cm}<{\centering} p{0.6cm}<{\centering}
p{0.6cm}<{\centering}
p{0.6cm}<{\centering} p{0.6cm}<{\centering}
}
\toprule
 & \multicolumn{6}{c}{CLEVR-Change} & \multicolumn{6}{c}{Spot-the-Diff} & \multicolumn{6}{c}{Editing-Request} \\
 &  \multicolumn{3}{c}{Image Pair $\Leftrightarrow$ Text} & \multicolumn{3}{c}{Text $\Leftrightarrow$ Image Pair} & \multicolumn{3}{c}{Image Pair $\Leftrightarrow$ Text} & \multicolumn{3}{c}{Text $\Leftrightarrow$ Image Pair} & \multicolumn{3}{c}{Image Pair $\Leftrightarrow$ Text} & \multicolumn{3}{c}{Text $\Leftrightarrow$ Image Pair} \\
Model &  R@1 &  R@5  &  R@10 & R@1 &  R@5  &  R@10 & R@10 &  R@20  &  R@50 & R@10 &  R@20  &  R@50  & R@1 &  R@5  &  R@10 & R@1 &  R@5  &  R@10  \\
\midrule
CLIP4IDC    &  46.4 & 83.0  &  86.6  &26.8  & 58.7  & 70.0 & 3.7  & 7.3  & 16.8   & 6.2  &  10.5  &  20.0 &  17.1  &  28.4  & 33.8  & 17.3  & 33.7 & 41.9  \\
\bottomrule
\end{tabular}}
\end{center}
\vspace{-4mm}
\caption{Results of IP-T and T-IP retrieval on the three datasets.  }
\label{table:retrieval_results}

\end{minipage}

\end{table*}

\begin{table*}[htbp]
\begin{minipage}{\linewidth}

\begin{center}

\scalebox{0.7}{\setlength\tabcolsep{6pt}
\begin{tabular}{p{1.5cm}
p{0.8cm}<{\centering}
p{0.8cm}<{\centering}
p{0.8cm}<{\centering}
p{0.8cm}<{\centering}
p{0.8cm}<{\centering} 
p{0.8cm}<{\centering} 
p{0.8cm}<{\centering}
p{0.8cm}<{\centering}
p{0.8cm}<{\centering}
p{0.8cm}<{\centering} 
p{0.8cm}<{\centering} 
p{0.8cm}<{\centering}
p{0.8cm}<{\centering}
p{0.8cm}<{\centering} a
p{0.8cm}<{\centering} 
p{0.8cm}<{\centering}
}
\toprule
 & & & \multicolumn{5}{c}{Image Pair $\Rightarrow$ Text} & \multicolumn{5}{c}{Text $\Rightarrow$ Image Pair} & \multicolumn{4}{c}{Captioning}\\
Model & $N_{intra}$ & $N_{inter}$ &  R@1 &  R@5  &  R@10  & MdR$\downarrow$  & MnR$\downarrow$ &  R@1 &  R@5  &  R@10  & MdR$\downarrow$  & MnR$\downarrow$ & B & M & C & R \\
\midrule
 \multirow{7}{*}{CLIP4IDC}    &  6 &  6 &  46.1  & 79.8  & 83.9  & 2.0   & 49.6 &  26.4  & 57.1  & 68.4  & 4.0   & 29.4  & 54.0 & 37.4 & 146.5 & 75.2 \\
    & 7  & 5  &  46.1  & \underline{80.8}  & \underline{84.5}  & \underline{2.0}   & \underline{45.5}  &  \underline{27.0}  & 57.8  & 69.0  & 4.0   & \underline{28.2}  & \underline{54.5} & \underline{37.5} & \underline{148.4} & \underline{75.5} \\
    & 8  & 4  &  \textbf{47.2}  & 80.7  & 84.4  &  2.0  & 46.3  &  \textbf{27.7}  & \textbf{58.7}  & \underline{69.7}  &  \underline{4.0}  & 29.9 & 54.1  & 37.4 &  147.3  &  75.4 \\
    &  9 & 3  &  \underline{46.4}  & \textbf{83.0}  & \textbf{86.6}  &  \textbf{2.0}  & \textbf{39.2}  & 26.8  & \underline{58.6}  & \textbf{70.0}  &  \textbf{4.0}  & \textbf{25.6} & \textbf{54.8}  & \textbf{37.8} & \textbf{148.6} & \textbf{75.8}  \\
  &  10 & 2  &  37.5  & 68.5  &  73.9 &  2.0  & 88.8  & 22.9  & 52.3  &  63.9 &  5.0  & 54.4  &  51.5 &  35.4  & 134.6  &  71.5 \\
    &  11 & 1  &  24.7  & 47.2  & 53.3  &  7.0  & 143.6 & 17.8  & 40.2  & 50.9  &  10.0  & 84.8  & 45.0 & 32.7 & 122.8 & 67.9 \\
    & 12  & 0  &  2.3  & 7.0  & 11.8 &  182.0  &  459.9  &  1.1  &  3.9  &  5.9  &  419.0  & 716.5  &  38.8 & 29.5  & 90.9  & 60.6  \\
\bottomrule
\end{tabular}}
\end{center}
\vspace{-4mm}
\caption{Results of setting different number of layers in CLIP4IDC on the IP-T, T-IP retrieval and IDC tasks on CLEVR-Change test split. }
\label{table:architectural_results}

\end{minipage}
\vspace{-2mm}
\end{table*}

\subsection{Adaptation Results}
We report the results in the retrieval tasks used for adaptation in Table~\ref{table:retrieval_results} on the test splits of the three datasets. These results from the image-pair and text retrieval tasks are simply to testify the model's capability of capturing details in the image pairs.
The effects brought by the retrieval tasks on the captioning accuracy are assessed in the following.

\section{Assessments of IDC Adaptation}

We study how the retrieval accuracy is affected by different architectural options in CLIP4IDC on CLEVR-Change test split.
Table~\ref{table:architectural_results} shows the effect of setting different numbers of layers in the \emph{intra} and \emph{inter} modules.
It can be seen that the improvement is achieved by allocating a large number of layers to the \emph{intra} module.
However, it does not mean that \emph{inter} layers are not required, as shown in the decreased accuracy when cutting the number of inter layers.
In addition, when the \emph{inter} layers are removed, i.e. $N_{inter}=0$, the architecture is similar to~\citet{luo2021clip4clip} and its accuracy is greatly reduced.
We owe it to the fact that the global information represented by two separate image embeddings fails to localize the changes between them.


To further study the relationships between the retrieval-based adaptation and the captioning accuracy, we fine-tune the models from the adaptation stage on the captioning task with the frozen image encoder. It can be observed in Table~\ref{table:architectural_results} that, in general, better adaptation with higher recall values on the retrieval tasks translates to better captioning. The observation suggests that the introduced retrieval tasks and the metrics used for retrieval serve as a strong indicator of the IDC performance.

\section{Conclusion and Future Work}

In this work, we studied how to fine-tune CLIP for image difference captioning. 
Retrieval-based adaptation was introduced to improve the visual representations for captioning and to narrow the gap between the purposes and data domains of CLIP pre-training and IDC.
Experimental results demonstrated the effectiveness of the CLIP4IDC model and the applied domain adaptation.

In the future work, we will further explore enhancing the relationships between the vision and language domains.
Specifically, 
CLIP4IDC adapts CLIP which does not involve cross-modal interactions as early as other pre-trained VL models \cite{lu2019vilbert,su2019vl,li2019visualbert} that allow the interactions from the ground up.
Adapting other VL models for IDC is naturally one interesting future direction. 
Moreover, exploring other means than our contrastive approach, such as domain confusion~\cite{tzeng2014deep}, to bridge vision and language domains is another plausible direction.

\section*{Acknowledgments}

This work has been supported by the Academy of Finland in projects 317388, 329268 and 345791. We also acknowledge the computational resources provided by both the Aalto Science-IT project and CSC -- IT Center for Science, Finland.

\bibliography{anthology,custom}

\appendix

\section{Dataset}
\label{sec:dataset}
CLEVR-Change~\cite{park2019robust} is a synthetic dataset generated by CLEVR engine.
Geometric differences between the objects in the images are annotated.
It is divided into the training, validation and test splits which have 67,660, 3,976 and 7,970 image pairs, respectively.
Spot-the-Diff~\cite{jhamtani2018learning} describes multiple scene changes in the real 13,192 image pairs sampled from the VIRAT Ground Video Dataset with human-annotated captions.
On an average, there are 1.86 sentences to describe the differences for each image pair.
Two decoding strategies containing single-sentence decoding and multi-sentence decoding are set for captioning.
Following~\citet{jhamtani2018learning}, we evaluate models in the single-sentence decoding by setting
the ground truth description as
multiple reference captions.
Image-Editing-Request~\cite{tan2019expressing} is a dataset consisted of camera shots, paintings and animations, and most of the images are realistic.
It contains 3,939 image pairs with instructions written by human annotators.

\section{Implementation Details}
\label{sec:implementation}
\noindent\textbf{IDC Adaptation Settings.}
The vision and language encoders are initialized with CLIP ViT-B/32~\cite{dosovitskiy2020image}.
The sentence length is 32 and the number of layers in the language encoder $N_G=$ 12.
The dimension of the text embedding $d_T=$ 512.
The size of an image is $224\times 224$ and each image is processed by a 2D convolution network with kernel size 32, stride 32 and 768 channels.
The number of image patches $n=$ 49 and the dimension of image patches $d_I=$ 768.
The number of layers in the intra- and inter-Transformer modules are $N_{intra}=9$ and $N_{inter}=3$, respectively. 
Adam optimizer is applied with initial learning rate $10^{-7}$.
The models are trained for 12 epochs by fixing all the random seeds to 42 on two NVIDIA Tesla V100 GPUs.

\noindent\textbf{IDC Fine-tuning Settings.}
We initialize the vision encoder with the model from IDC adaptation and set the dimensionality of the word embedding $d_T=512$.
The captioning model is learned from scratch.
The number of Transformer layers in both captioning encoder and decoder is 3 on all the datasets. 
The attention layer in the Transformer has 8 heads and 10\% dropout probability, and its hidden size is 512.

For the direct CLIP fine-tuning, the parameters of its vision encoder are initialized with CLIP ViT-B/32.
The settings of its captioning model are the same as those in CLIP4IDC.

Adam is used with initial learning rate $10^{-7}$ for the vision encoder and $10^{-4}$ for the captioning model.
The model is trained for at most 50 epochs and the batch size is 16.
Greedy decoding with maximum 32 steps is applied for generating sentences in inference.
The experiments are carried out on a NVIDIA Tesla V100 GPU.

\section{Qualitative Results}
\label{sec:qualitative}

\begin{figure*}[htbp]

\begin{minipage}{0.45\linewidth}
\begin{subfigure}[b]{\linewidth}
\centering
\captionsetup{justification=centering}
\includegraphics[width=0.99\linewidth]{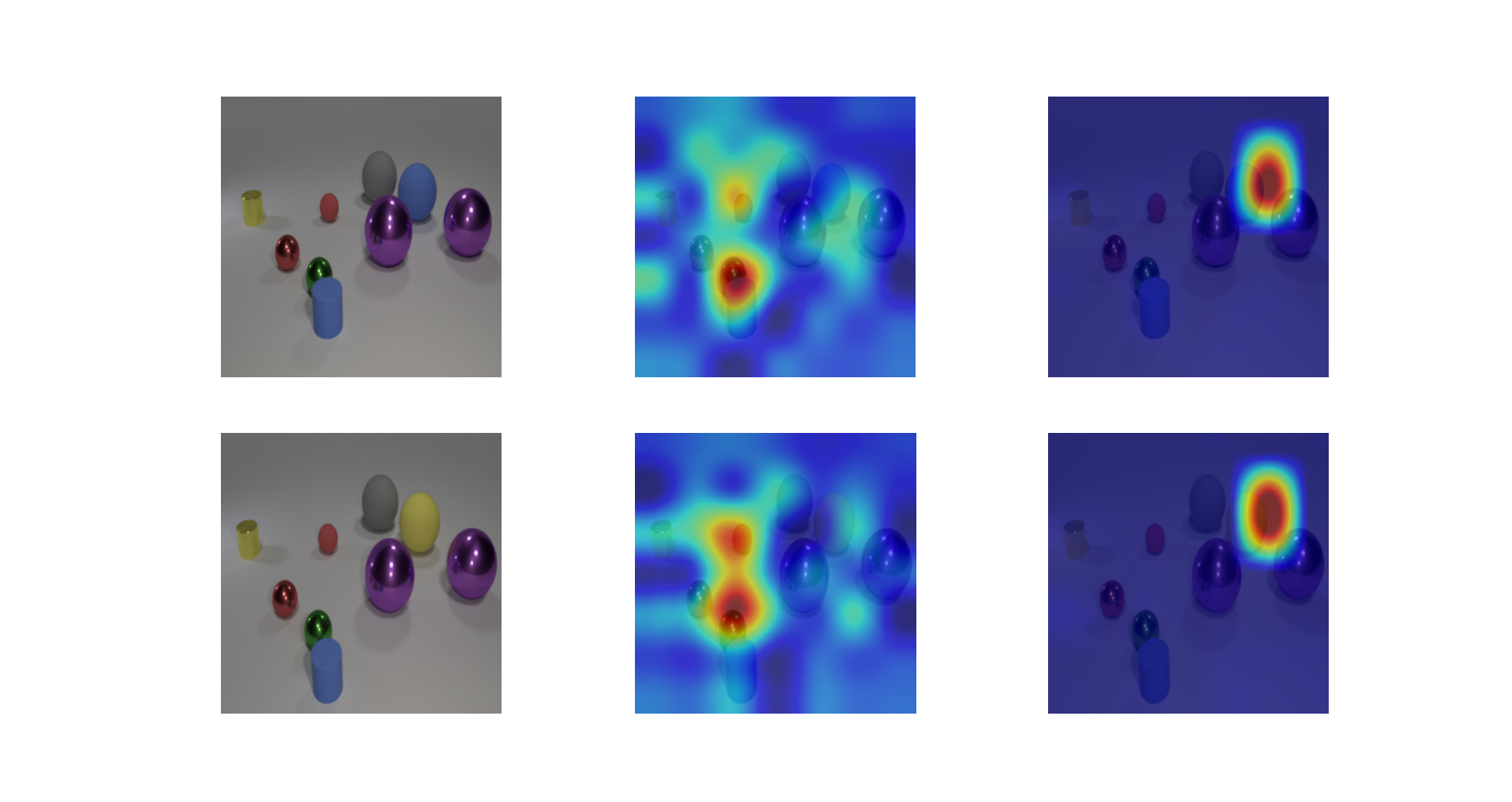}
\caption*{GT: the blue ball changed to yellow\\
CLIP4IDC: the blue ball became yellow}
\end{subfigure}

\end{minipage}
\begin{minipage}{0.45\linewidth}

\begin{subfigure}[b]{\linewidth}
\centering
\captionsetup{justification=centering}
\includegraphics[width=0.99\linewidth]{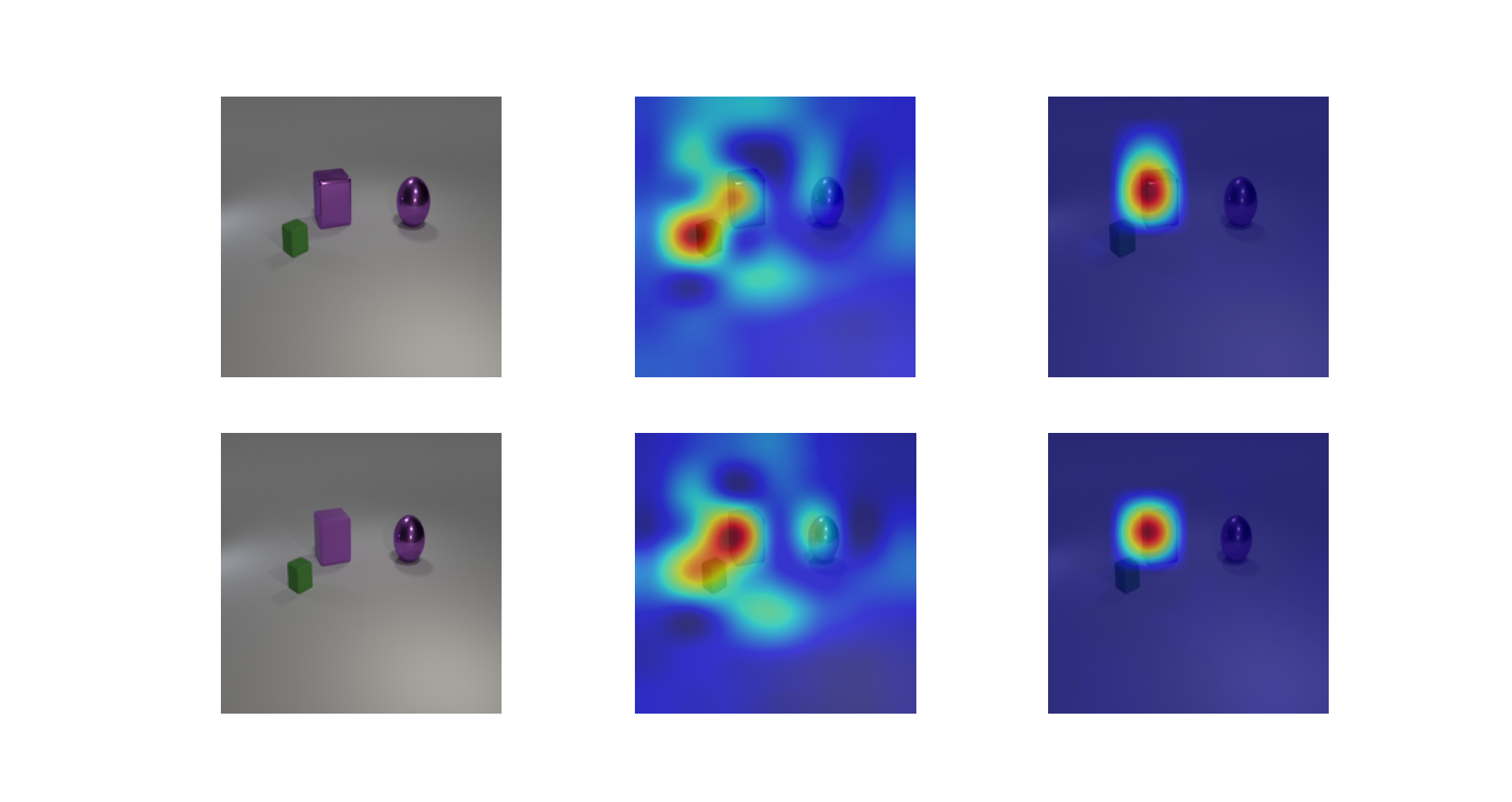}
\caption*{GT: the big purple metal block behind the green thing changed to rubber\\
CLIP4IDC: the large purple metal block that is behind the big purple metal sphere became rubber}
\end{subfigure}

\end{minipage}

\begin{minipage}{0.45\linewidth}

\begin{subfigure}[b]{\linewidth}
\centering
\captionsetup{justification=centering}
\includegraphics[width=0.99\linewidth]{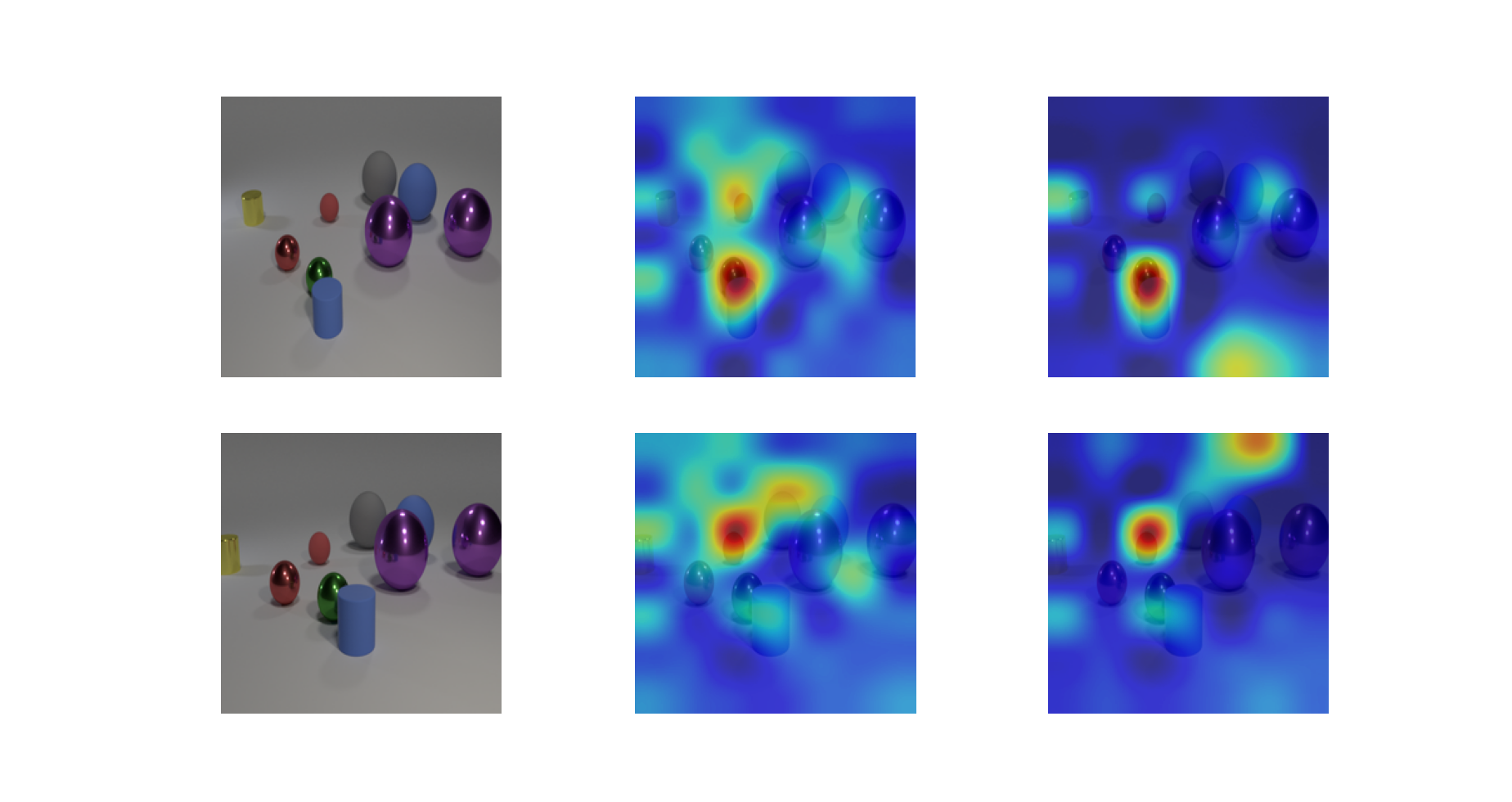}
\caption*{GT: there is no difference\\
CLIP4IDC: there is no change}
\end{subfigure}

\end{minipage}
\begin{minipage}{0.45\linewidth}

\begin{subfigure}[b]{\linewidth}
\centering
\captionsetup{justification=centering}
\includegraphics[width=0.99\linewidth]{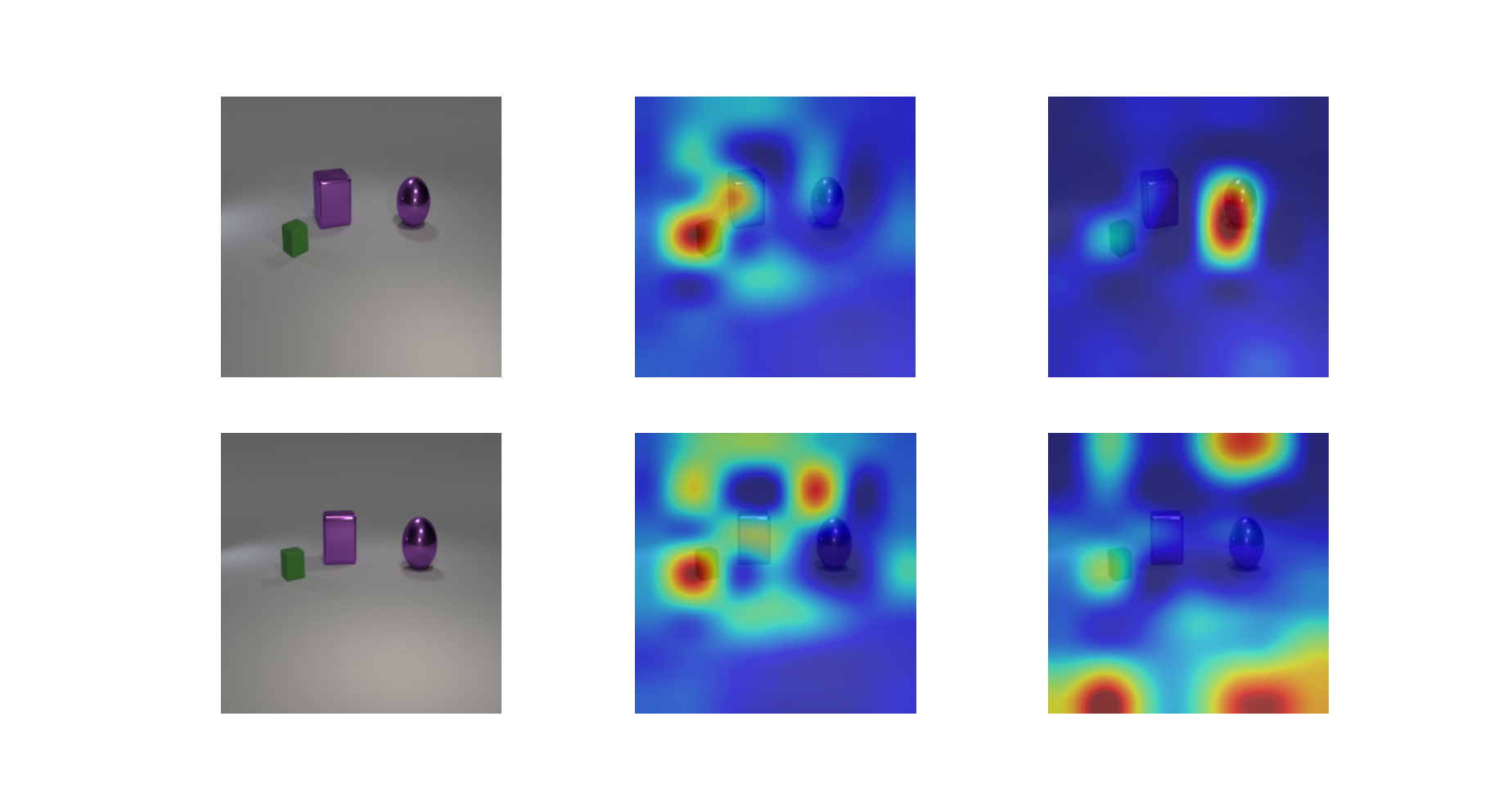}
\caption*{GT: there is no change\\
CLIP4IDC: there is no change}
\end{subfigure}

\end{minipage}

\caption{Visualization of the vision encoder's output in CLIP4IDC on CLEVR-Change. Figures are arranged in three columns. The first column shows the first and the second raw images. The second column shows their attention maps in the intra-encoder's output. The last column shows their attention maps in the inter-encoder's output.
}
\label{figure:clevr}

\end{figure*}

\begin{figure*}[htbp]

\begin{minipage}{0.45\linewidth}

\begin{subfigure}[b]{\linewidth}
\centering
\captionsetup{justification=centering}
\includegraphics[width=0.99\linewidth]{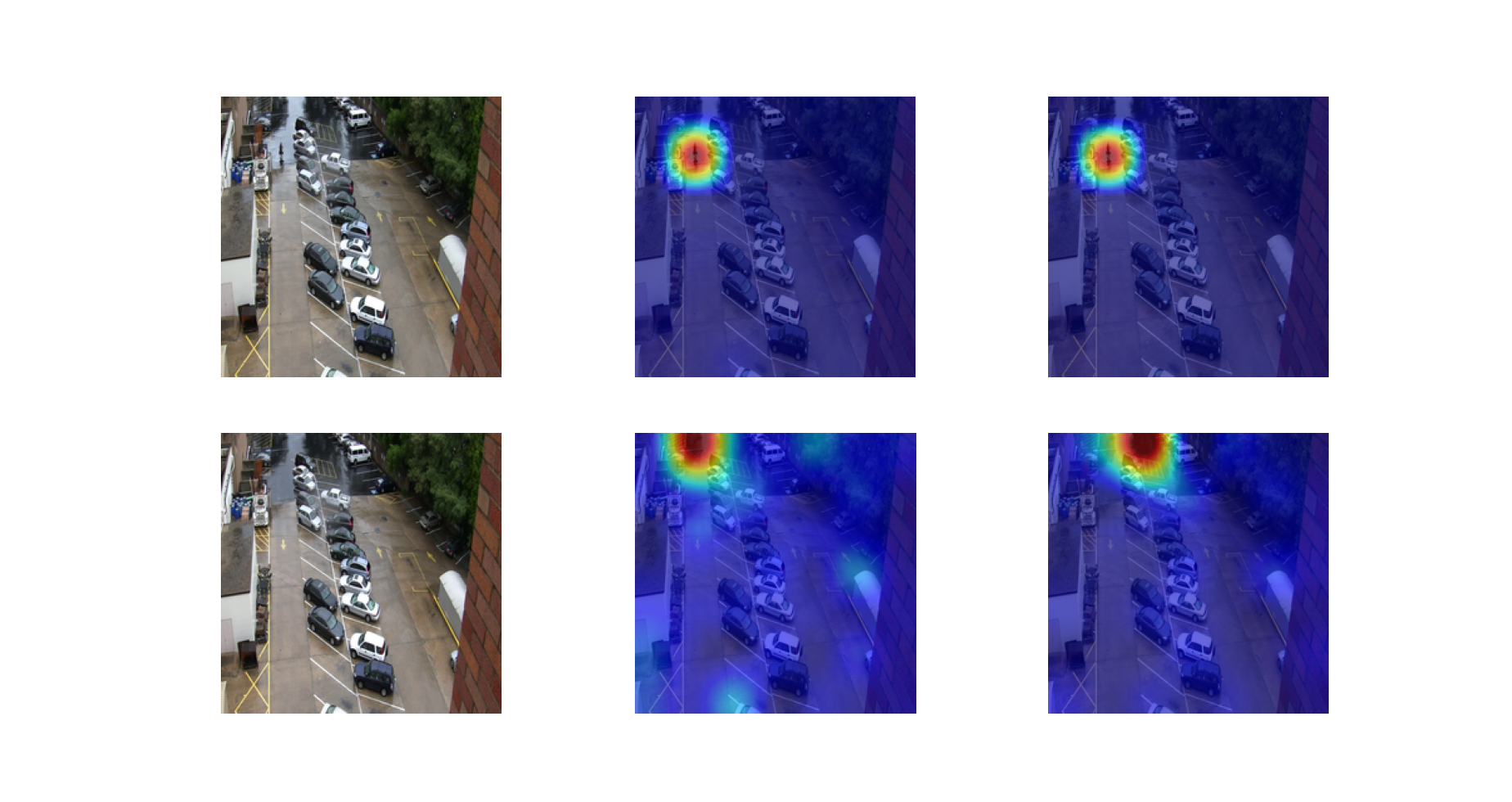}
\caption*{
GT: the person walking is no longer there \\
CLIP4IDC: the person walking in the parking lot is gone}
\end{subfigure}
\end{minipage}
\begin{minipage}{0.45\linewidth}

\begin{subfigure}[b]{\linewidth}
\centering
\captionsetup{justification=centering}
\includegraphics[width=0.99\linewidth]{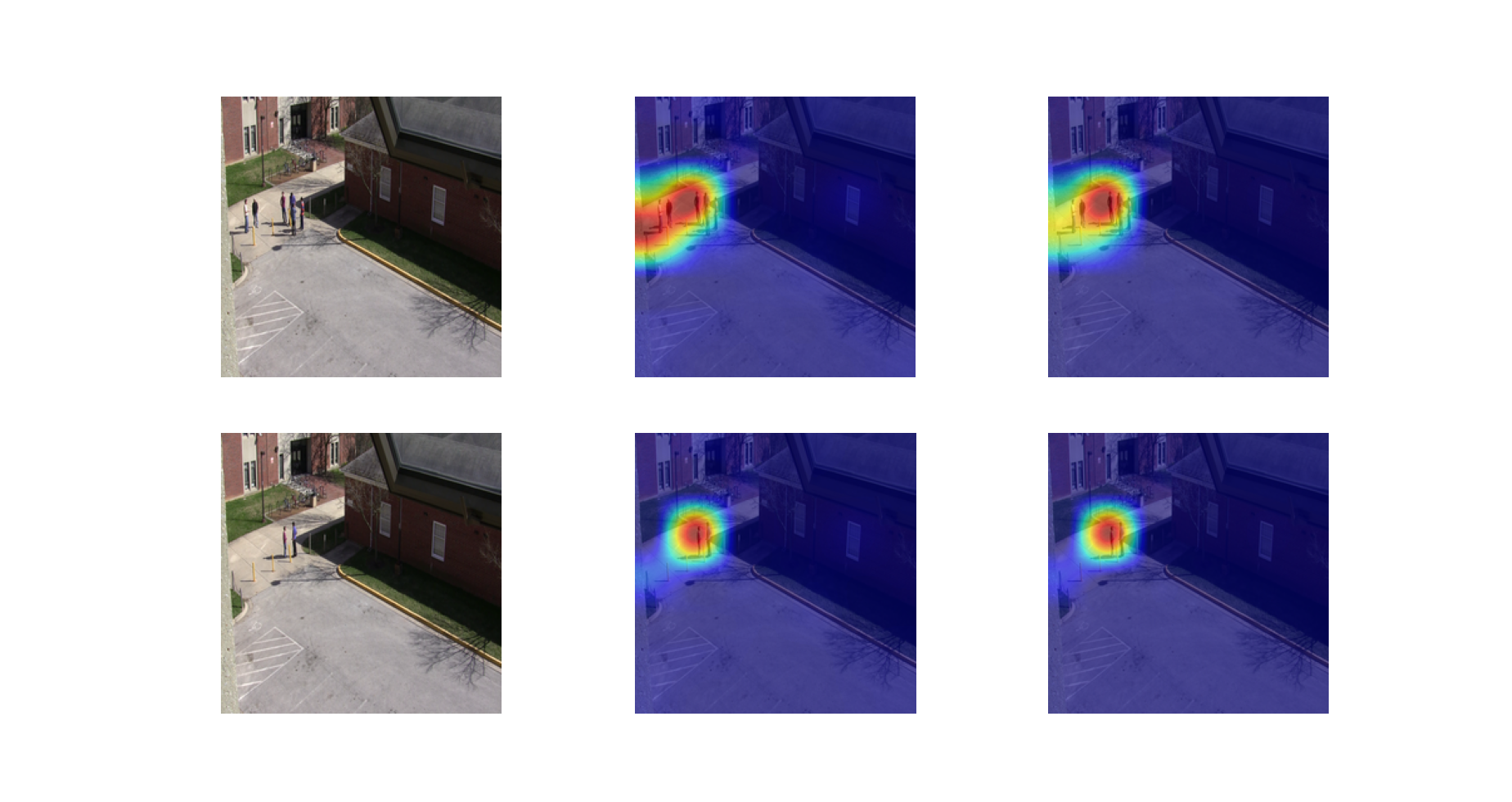}
\caption*{
GT: there is a smaller group of people in the lot \\
CLIP4IDC: there are two people in the right image}
\end{subfigure}

\end{minipage}

\begin{minipage}{0.45\linewidth}

\begin{subfigure}[b]{\linewidth}
\centering
\captionsetup{justification=centering}
\includegraphics[width=0.99\linewidth]{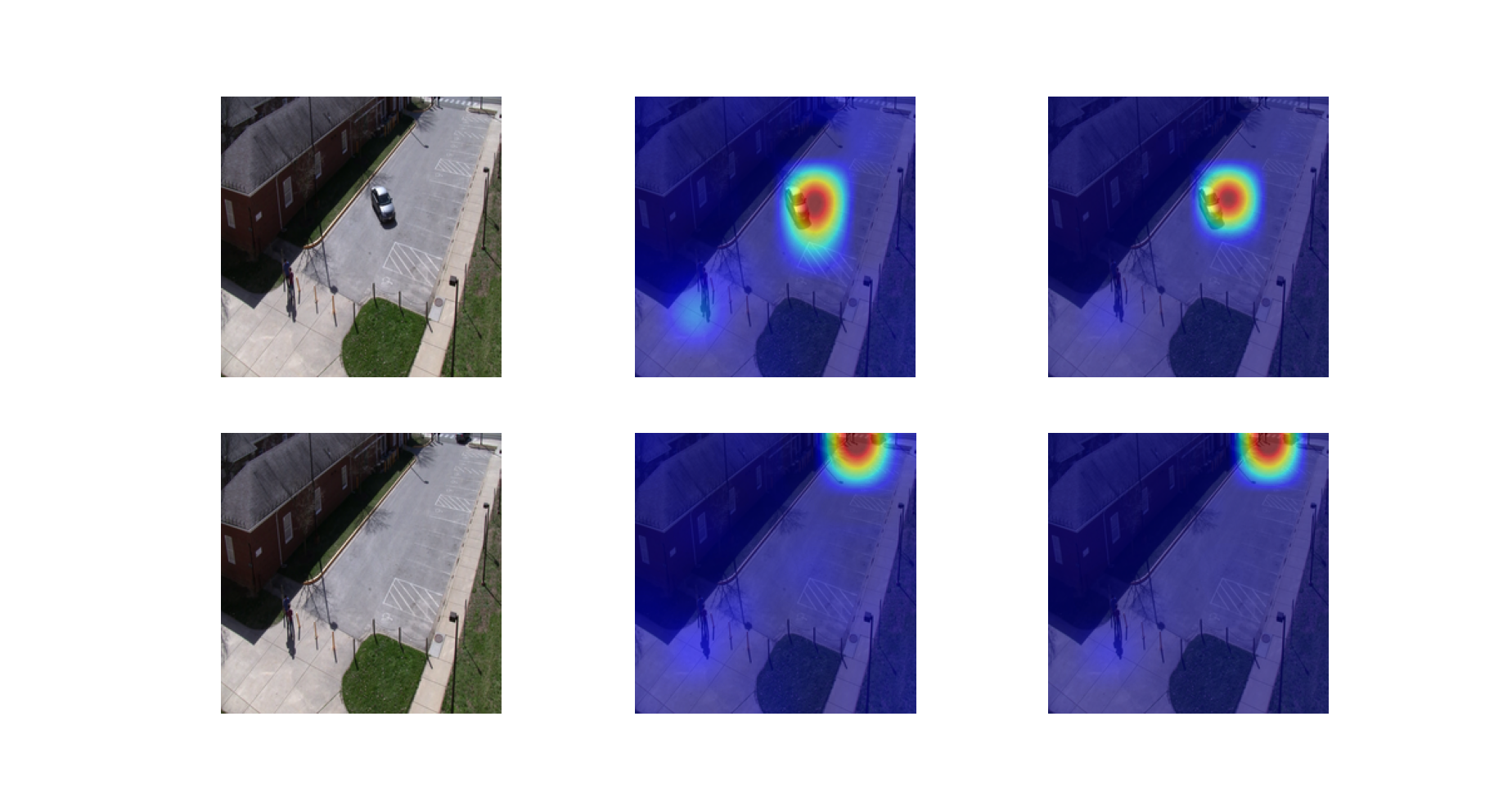}
\caption*{
GT1: the car is gone\\
GT2: there is a car entering from the entrance at the top right of the image \\
CLIP4IDC: the car is gone}
\end{subfigure}

\end{minipage}
\begin{minipage}{0.45\linewidth}

\begin{subfigure}[b]{\linewidth}
\centering
\captionsetup{justification=centering}
\includegraphics[width=0.99\linewidth]{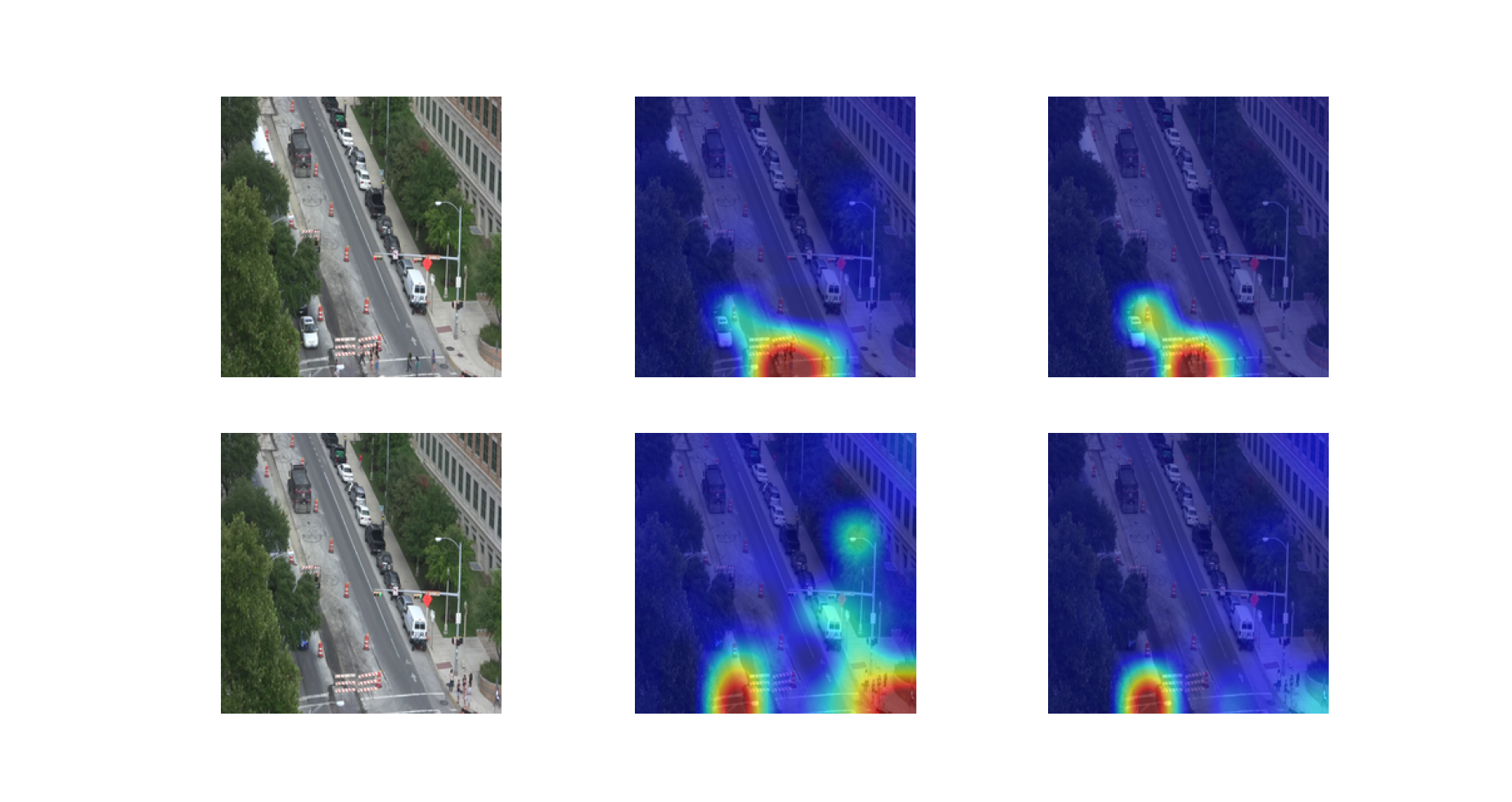}
\caption*{
GT1: the white car in the left corner is gone\\
GT2: there are now people waiting to cross the intersection \\
CLIP4IDC: there are people walking on the sidewalk}
\end{subfigure}
\end{minipage}

\caption{Visualization of the vision encoder's output in CLIP4IDC on Spot-the-Diff. }
\label{figure:spot}

\end{figure*}

\begin{figure*}[htbp]

\begin{minipage}{0.45\linewidth}

\begin{subfigure}[t]{\linewidth}
\centering
\captionsetup{justification=centering}
\includegraphics[width=0.99\linewidth]{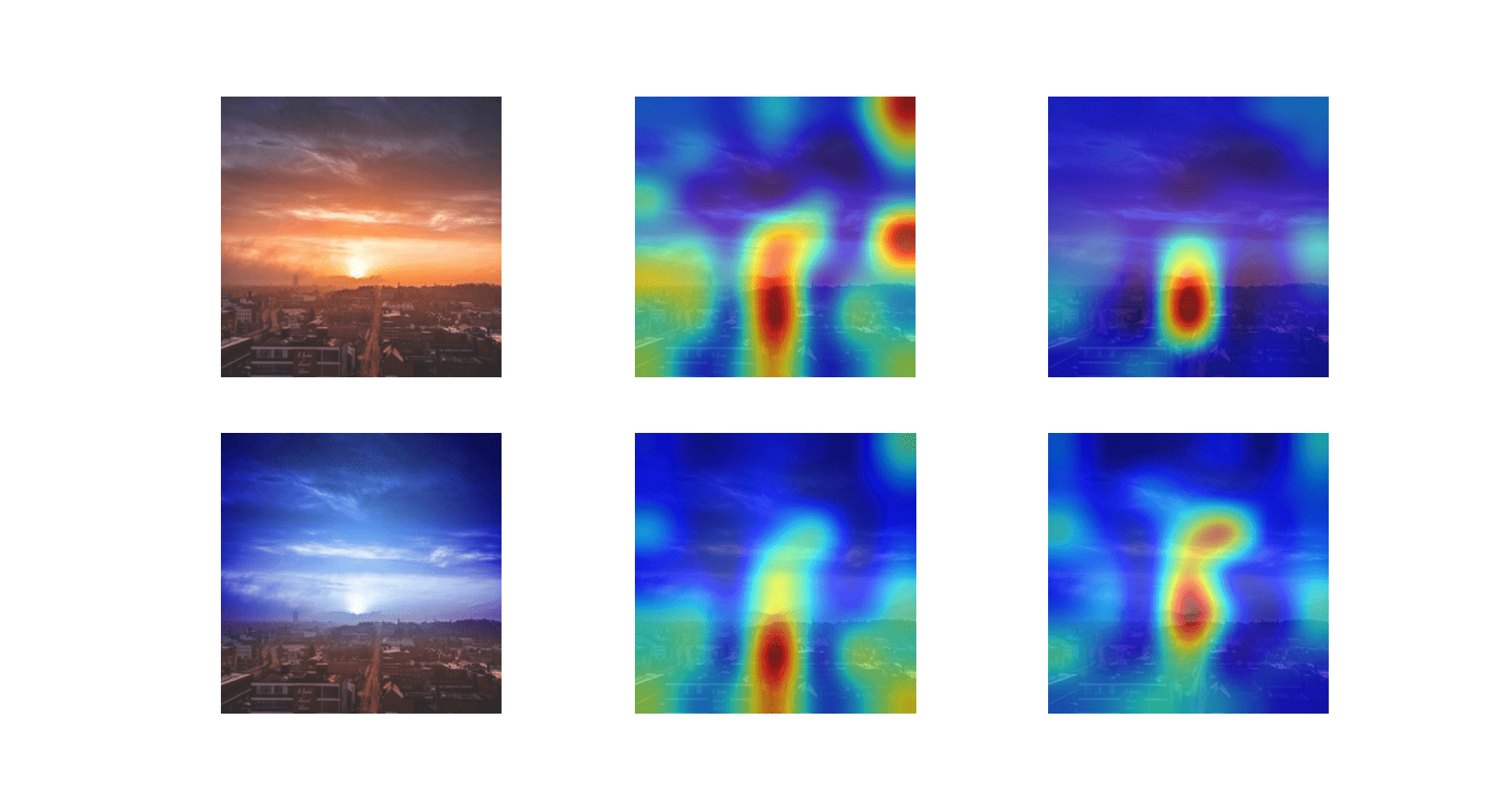}
\caption*{
GT: color the sky blue\\
CLIP4IDC: make the image more blue}
\end{subfigure}
\end{minipage}
\begin{minipage}{0.45\linewidth}

\begin{subfigure}[t]{\linewidth}
\centering
\captionsetup{justification=centering}
\includegraphics[width=0.99\linewidth]{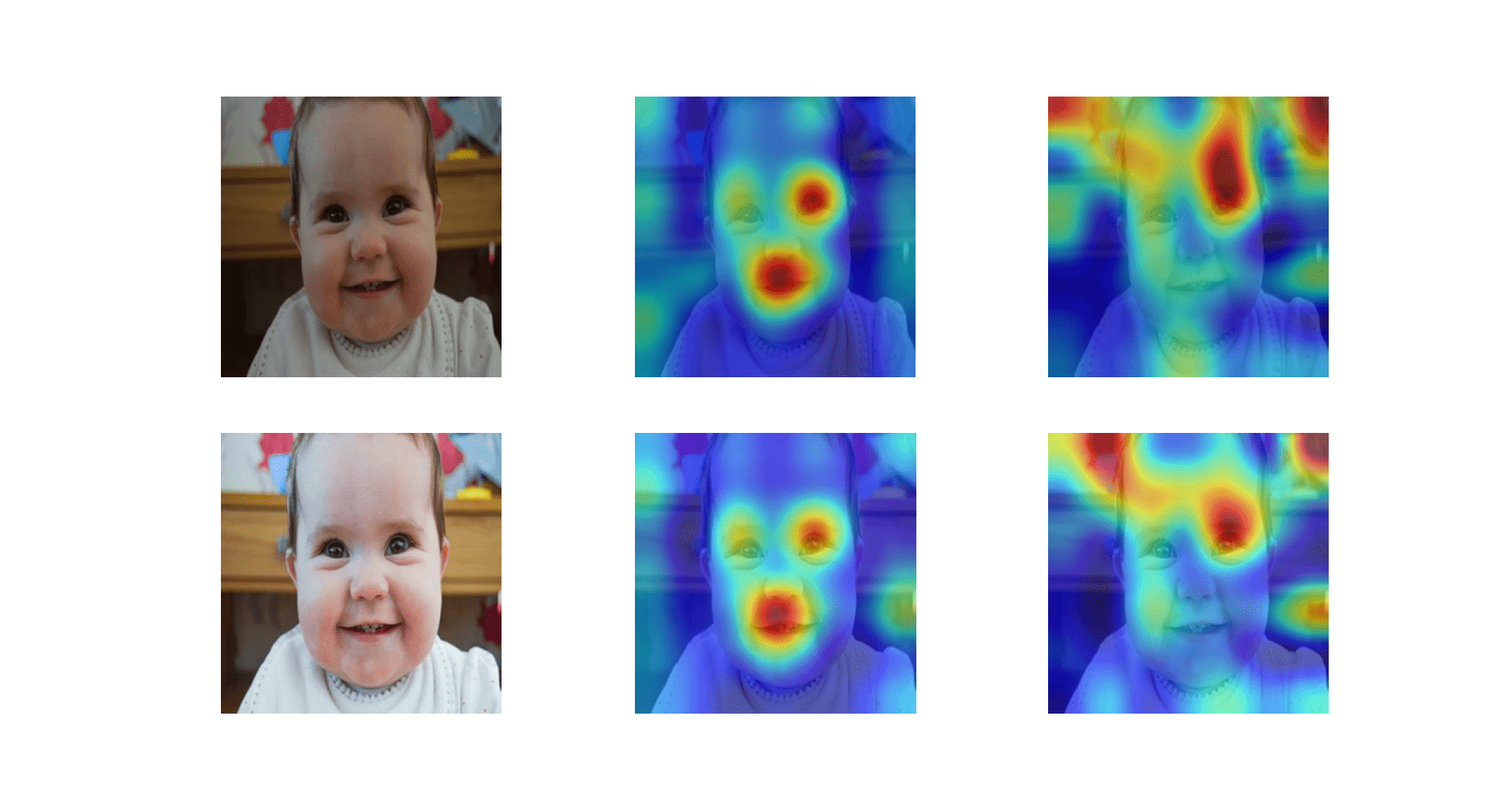}
\caption*{
GT: brighten the entire photo\\
CLIP4IDC: brighten the photo}
\end{subfigure}

\end{minipage}

\begin{minipage}{0.45\linewidth}

\begin{subfigure}[t]{\linewidth}
\centering
\captionsetup{justification=centering}
\includegraphics[width=0.99\linewidth]{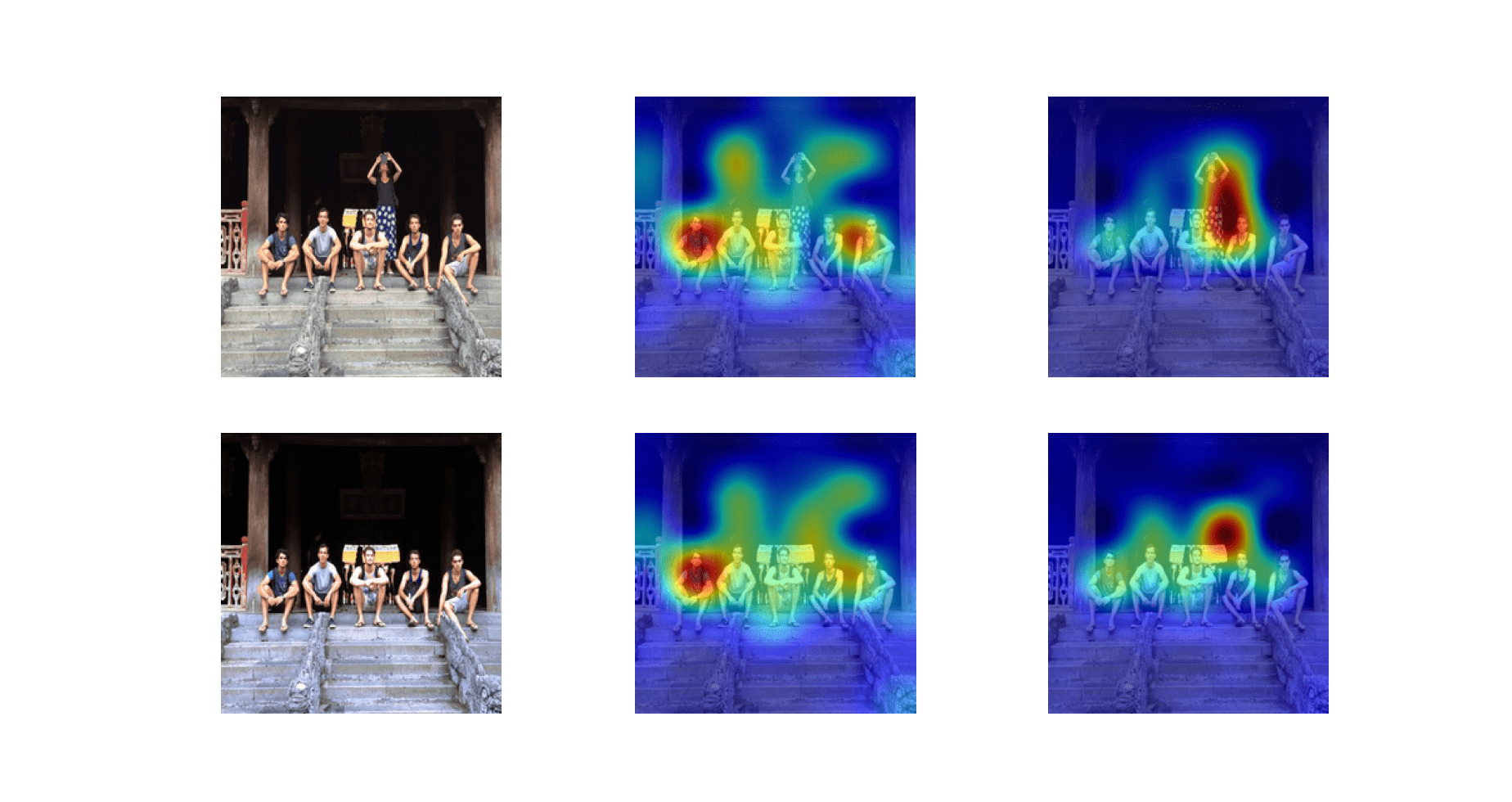}
\caption*{
GT: remove girl in background\\
CLIP4IDC: remove the people from the background}
\end{subfigure}

\end{minipage}
\begin{minipage}{0.45\linewidth}

\begin{subfigure}[t]{\linewidth}
\centering
\captionsetup{justification=centering}
\includegraphics[width=0.99\linewidth]{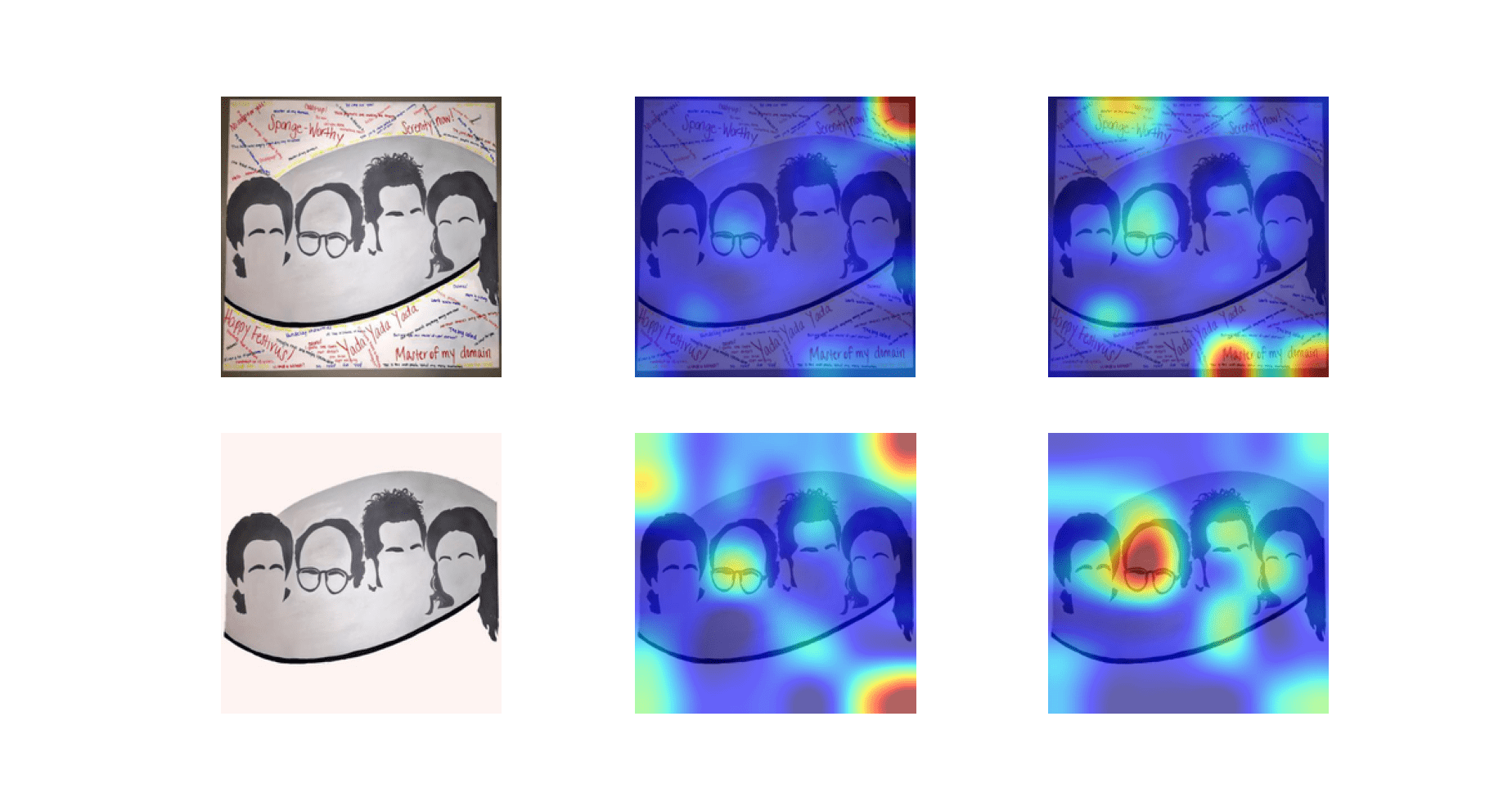}
\caption*{
GT: remove the background\\
CLIP4IDC: remove the background}
\end{subfigure}
\end{minipage}

\caption{Visualization of the vision encoder's output in CLIP4IDC on Image-Editting-Request. }
\label{figure:edit}

\end{figure*}

To understand the effect of IDC adaptation, some cases on CLEVR-Change, Spot-the-Diff and Image-Editting-Request datasets are visualized in Figures~\ref{figure:clevr},~\ref{figure:spot} and~\ref{figure:edit}, respectively.

\paragraph{Synthetic Dataset}
The four cases in Figure~\ref{figure:clevr} are from CLEVR-Change. 
In the second column of each case, it can be seen that CLIP4IDC's \emph{intra} encoder attends to regions where information is more likely to be needed for capturing the fine-grained difference in the second images.
While in the third column of them, \emph{inter} encoder filters the information uncorrelated to the difference and pay attention to the changes in the second image.
However, the condition is different for the cases, shown in the bottom two sets of figures, without changes.
The \emph{inter} encoder appears to attend more uniformly across regions to seek for any change instead of getting fixated on one specific region.

\paragraph{Real-world Dataset}
Figures~\ref{figure:spot} and~\ref{figure:edit} show the cases from Spot-the-Diff and Image-Editing-Request, respectively.
It can be seen that our CLIP4IDC capture the fine-grained differences in the real-world and complicated cases.

\section{Descriptions of the Baseline Methods}
\label{sec:related_work}

Some recent works have made great progress in the IDC task by devising a language model that describes the changes, given the visual features pre-extracted by the CNN backbones~\cite{he2016deep,ren2015faster}. We describe the baselines we compare against in the experiments as follows:

\begin{itemize}[itemsep=0pt,itemsep=-1ex]
\item \textbf{DUDA}~\citeyearpar{park2019robust}: A dual attention module is proposed to distinguish distractors from semantic changes and localize the changes. A dynamic attention module is then used to describe the changes.
\item \textbf{VAM}~\citeyearpar{shi2020finding}: A novel visual encoder is proposed to distinguish viewpoint changes from semantic changes.
Moreover, it fine-tunes the model directly with reinforcement learning in which the rewards coming from evaluating the generated captions.

\item \textbf{IFDC}~\citeyearpar{huang2021image}: A language generator, which consists of a feature fusion module, a similarity-based difference finding module, and a difference captioning module, is introduced.
\item \textbf{VACC}~\citeyearpar{kim2021agnostic}: A difference encoder is devised to encode viewpoint information and model the difference.
\item \textbf{BiDiff}~\citeyearpar{sun2022bidirectional}: A change captioning pipeline is introduced to localize the changes in the image pair and a decoder with spatial-channel attention to generate descriptions.
\end{itemize}

These methods consistently improve the model accuracy by refining or improving the visual features to better capture the fine-grained changes in the image pair.
In addition, inspired by the success of multi-task learning, the following training schemes were also introduced.

\begin{itemize}[itemsep=0pt,itemsep=-1ex]
\item \textbf{VACC}~\citeyearpar{kim2021agnostic} and \textbf{DUDA+Aux}~\citeyearpar{hosseinzadeh2021image}: Both work proposed auxiliary modules to match the composite feature of the generated caption and before image with the after image feature.
\item \textbf{IDC-PCL}~\citeyearpar{yao2022image}: A "pretrain-and-finetune" paradigm is proposed and contains three pretraining tasks as follows.
Given visual-linguistic contexts, the Masked Language Modelling (MLM) and Masked Visual Contrastive Learning (MVCL) tasks were applied to map the visual context to language and to reconstruct the masked image features, respectively.
Fine-grained Difference Aligning (FDA) was introduced to rewrite the captions as the hard samples to maximize the connections in the joint representation of the text and the image pair.
\item \textbf{CC-Full}~\citeyearpar{ak2022learning}: The work proposed to co-train text-based image manipulation (TIM) with change captioning (CC) modules. 
The CC module generates captions evaluated with the TIM module with a reinforcement learning framework.
The TIM module generates images that are evaluated with the CC module with a generative adversarial network.
\end{itemize}

\end{document}